\pdfoutput=1

\documentclass[11pt]{article}

\usepackage{acl}

\usepackage{times}
\usepackage{latexsym}

\usepackage[T1]{fontenc}

\usepackage[utf8]{inputenc}

\usepackage{microtype}

\usepackage{inconsolata}

\usepackage{graphicx}

\usepackage{tabularray}
\usepackage{url}
\usepackage{twemojis}
\usepackage{soul}
\soulregister\citep7

\newcommand{\yes}{\scalebox{1.25}{\twemoji{check mark button}}}
\newcommand{\no}{\scalebox{1.25}{\twemoji{cross mark}}}
\newcommand{\pone}{\scalebox{1.2}{\twemoji{keycap: 1}}}
\newcommand{\ptwo}{\scalebox{1.2}{\twemoji{keycap: 2}}}
\newcommand{\pthree}{\scalebox{1.2}{\twemoji{keycap: 3}}}
\newcommand{\pfour}{\scalebox{1.2}{\twemoji{keycap: 4}}}
\newcommand{\pfive}{\scalebox{1.2}{\twemoji{keycap: 5}}}
\newcommand{\psix}{\scalebox{1.2}{\twemoji{keycap: 6}}}
\newcommand{\pseven}{\scalebox{1.2}{\twemoji{keycap: 7}}}
\newcommand{\peight}{\scalebox{1.2}{\twemoji{keycap: 8}}}
\newcommand{\pnine}{\scalebox{1.2}{\twemoji{keycap: 9}}}
\newcommand{\fdir}{\scalebox{1.2}{\twemoji{triangular flag}}}
\newcommand{\gear}{\scalebox{1.2}{\twemoji{gear}}}

\newcommand{\mg}[1]{{\textcolor{black}{#1}}}
\newcommand{\mn}[1]{{\textcolor{black}{#1}}}
\newcommand{\sara}[1]{{\textcolor{black}{#1}}}
\newcommand{\lb}[1]{{\textcolor{black}{#1}}}

\colorlet{importcolor}{yellow!30}
\sethlcolor{importcolor}

\title{\mg{Speech Translation with Speech Foundation Models and\\ Large Language Models: What is There and What is Missing?}}

\author{Marco Gaido \and Sara Papi \and Matteo Negri \and Luisa Bentivogli \\
  Fondazione Bruno Kessler, Trento, Italy \\
  \texttt{\{mgaido,spapi,negri,bentivo\}@fbk.eu} \\}

\begin{document}
\maketitle
\begin{abstract}
The field of natural language processing (NLP) has recently witnessed a transformative shift with the emergence of foundation models, particularly Large Language Models (LLMs) that have revolutionized text-based NLP. This paradigm has extended to other modalities, including speech, where researchers \mg{are} actively
\mg{exploring}
the combination of Speech Foundation Models (SFMs) and LLMs 
\lb{into single, unified models}
capable of addressing multimodal tasks. Among such tasks, this paper focuses on speech-to-text translation (ST). 
By examining the published papers on the topic, 
we propose a unified view of the architectural solutions and training strategies presented so far, highlighting similarities and differences among them.
Based on this examination, \mg{we \sara{not only} 
\mn{organize}
the lessons learned but also 
\lb{show} how diverse settings and evaluation approaches hinder the identification of the best-performing solution for each architectural building block and training choice.} 
Lastly, we \sara{outline} recommendations for future works on the topic aimed at better understanding the strengths and weaknesses of the 
\sara{SFM+LLM solutions for ST.}
\end{abstract}

\section{Introduction}



The natural language processing (NLP) landscape has recently undergone a 
\lb{paradigm shift} with the emergence of foundation models \citep{Bommasani2021FoundationModels}. Among them, Large Language Models (LLMs) have revolutionized text-based NLP, showcasing remarkable capabilities across a wide range of NLP tasks \citep{radford2019language}. This unprecedented success has spurred research into creating foundation models for other modalities, including speech processing \citep{latif2023sparks}.


\begin{figure*}[!t]
    \centering
    \includegraphics[width=0.9\textwidth]{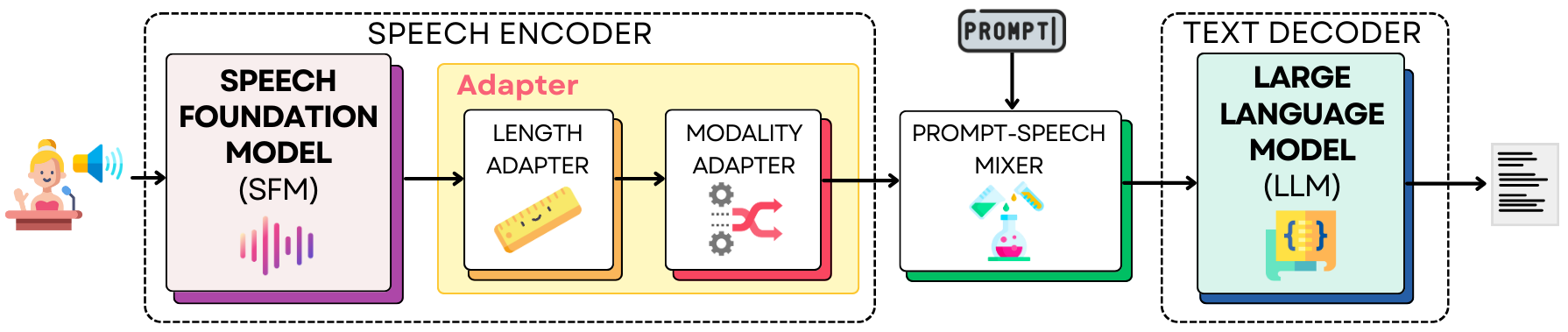}
    \caption{\mg{Architectural building blocks of ST models based on the combination of an SFM and an LLM}.}
    \label{fig:speech_and_llm}
\end{figure*}

Building on the translation abilities of LLMs \citep{hendy2023good,jiao2023chatgpt,raunak-etal-2023-gpts,zhu2023multilingual,xu2023paradigm} and the 
remarkable speech recognition and understanding capabilities
achieved by Speech Foundation Models (SFMs) \citep{pmlr-v202-radford23a,pratap2023mms,communication2023seamlessm4t}, researchers are now actively exploring 
\mg{their combination. The resulting large multimodal models
leverage, on \sara{the} one \sara{hand}, the SFM ability to encode speech content into rich and high-level representations and, on the other, the extensive linguistic knowledge of the LLM to}
\mg{generate fluent outputs and address a wide range of tasks}
\citep{chen2023x,yu2023connecting,wang2023speechtotext,rubenstein2023audiopalm,zhang-etal-2023-speechgpt}.
Focusing on the
speech-to-text translation (ST) task
-- the scope of this paper -- 
the rapid pace of the advancements has led to multiple parallel  
endeavors, resulting in a variety of solutions.
While all these efforts have the merit of demonstrating the viability and effectiveness of this 
\mn{line of work},
their contemporaneity, along with methodological inconsistencies, 
hinders
\mg{a fair}
comparison\mg{.}
\mn{For this reason, 
we provide a
systematic analysis of the proposed SFM+LLM solutions \mg{for ST} with the multiple goals of identifying \sara{their} similarities and differences, organizing the lessons learned, and suggesting future research directions, along with best practices for insightful evaluations.}
%
%
%
\mn{At its core, this paper addresses two key questions:}

\begin{itemize}
    \item[\fdir] \textbf{What is There?} We survey the publicly available works that propose
    \mg{an SFM+LLM solution for ST,}
    resulting in 9 papers (henceforth referred to as \pone{},...,\pnine{}), and analyze them ($\S$\ref{sec:there}) focusing on two orthogonal aspects:
    \begin{itemize}
        \item[\gear] \textbf{Architectural Building Blocks} 
        ($\S$\ref{subsec:architecture}): We delve into the SFM+LLM \mg{architectures,}
        \mn{identifying} a common abstraction made of 5 building blocks and
        underscoring similarities
        \mn{and differences} 
        in 
        the 
        SFM and LLM choices,
        \sara{along with}
        the strategies adopted for combining them\mg{;}
        \item[\gear] \textbf{Training and Evaluation} 
        ($\S$\ref{subsec:train-eval}):
        We inspect the \mg{training data, tasks, and strategies employed in the studies, \sara{as well as evaluation data and supported language pairs,} gathering insights \sara{about} promising solutions,} 
        and \sara{highlighting} the sparsity of the current landscape;
    \end{itemize}
    \item[\fdir] \textbf{What is Missing?}  
    \mg{We conclude by underscoring the \sara{importance} 
    \mn{of establishing a standard training setting}
     based on open data
    \sara{to 
    \mn{ease}
    direct comparability across works,}
    and \mn{by} identifying aspects that need further investigation to better understand the potential of}
    \mn{SFM+LLM combination for ST}
    ($\S$\ref{sec:missing}).
\end{itemize}

\begin{table*}[!t]
\small
    \centering
        \begin{tblr}{
      colspec={|X[0.2,c]|X[1.5]|X[1.5]|X[1.5]|X[1.5]|X[1.5]|X|X|X}, row{1} = {c}, hlines,
      cell{3}{4}={c=2}{},
      hspan=minimal,
    }
        \# & \textbf{Model} & \textbf{SFM} & \textbf{LA} & \textbf{MA} & \textbf{LLM} & \textbf{Prompt} & \textbf{PSMix} \\
        \hline
        \pone & LST \citep{zhang2023tuning} & wav2vec 2.0 \citep{NEURIPS2020_92d1e1eb} & 2$\times$Conv1D & 1 FFN & LLaMa2 13B \citep{touvron2023llama} & None & Speech Only \\
        \ptwo & SALM \citep{chen2023salm} & NeMo STT Fast Conformer \cite{stt_nvidia} & 2$\times$Conformer layers with 4$\times$ downsample & & Megatron-LM 2B \citep{shoeybi2020megatronlm} & Fixed Template & Speech Prepended \\
        \pthree{} & Speech-LLaMa \citep{wu2023decoder} & \SetCell[r=2]{c} in-house Transformer & \SetCell[r=2]{c} CTC compression \citep{gaido-etal-2021-ctc} & \SetCell[r=2]{c} 4 Transformer Layers + 1 FFN & \SetCell[r=2]{c} LLaMa2 7B \citep{touvron2023llama} & Sampled from List of Templates & Speech Appended \\
        \pfour & COSMIC \citep{pan2023cosmic} & & & & & Fixed for ASR/ST, Open for SQA & Speech Prepended \\
        \pfive & SLM \citep{wang2023slm} & USM \citep{zhang2023google} & Randomly discarded 75\% vectors & 2 Transformer Layers & mT0-MT XXL 13B \citep{muennighoff-etal-2023-crosslingual} & Fixed for ASR/ST, Open for SIT & Speech Appended \\
        \psix & SALMONN \citep{tang2023salmonn} & Whisper-large-v2 \citep{pmlr-v202-radford23a} + BEATs \citep{pmlr-v202-chen23ag} & \SetCell[c=2]{c}{Window-level Q-Former \citep{10.5555/3618408.3619222}} & & Vicuna 13B \citep{vicuna2023} & Fixed for ASR/ST, Open for Other Tasks & Speech Prepended \\
        \pseven & LLM-ST \citep{huang2023speech} & Whisper-large-v3 \citep{pmlr-v202-radford23a} & NA & NA & GPT 13B \citep{NEURIPS2020_1457c0d6} trained from scratch & Fixed & Speech Prepended \\
        \peight & Qwen-Audio \citep{chu2023qwen} & Whisper-large-v2 \citep{pmlr-v202-radford23a} & NA & NA & Qwen 7B \citep{bai2023qwen} & Learned Tokens & Speech Prepended \\
        \pnine & Conformer LLaMa \citep{fathullah2023towards} & Custom Conformer trained on ASR data & Stacking 4 consecutive vectors & 1 FFN & LLaMa2 Chat 7B \citep{touvron2023llama} & LLaMa's Default Structure & Speech Placed within the Prompt \\
        \end{tblr}%
    \caption{Architectural components of SFM+LLM comprising speech foundation model (SFM), length adapter (LA), modality adapter (MA), large language model (LLM), prompt, and prompt-speech mixer (PSMix).}
    \label{tab:speech_llm_components}
\end{table*}

\section{What is There?}
\label{sec:there}

In this section, we explore two key aspects of SFM+LLM research \mn{in ST}: first, we delve into the architectural components of SFM+LLM models
($\S$\ref{subsec:architecture});
second, we examine the training and evaluation settings utilized in these studies
($\S$\ref{subsec:train-eval}).

\subsection{Architectural Building Blocks}
\label{subsec:architecture}

The combination of SFMs and LLMs
has so far been addressed 
\mn{with} different architectures, which have, though, a common structure. Specifically, we identify 5 building blocks (see Figure \ref{fig:speech_and_llm}): \textit{i)} the SFM, \textit{ii)} the length adapter, \textit{iii)} the modality adapter, \textit{iv)} the prompt-speech mixer that merges the textual prompt with the adapted speech representation, and \textit{v)} the LLM. In Table \ref{tab:speech_llm_components}, we summarize how the 9 analyzed papers have designed each component.

\paragraph{SFM.} 
The SFM is in charge of extracting
rich, semantic representations from the audio signal, which have then to be projected onto the LLM input semantic space to successfully connect the audio modality with the LLM. Looking at Table \ref{tab:speech_llm_components}, we immediately notice that there is no consensus on the best SFM to choose. 
With the only exception of \pthree{} and \pfour, which are from the same authors/research group, each work relies on a different SFM. Also,
no work has 
addressed the comparative assessment of different SFMs under controlled conditions within the same framework.
Differences among SFMs encompass multiple aspects. First, their architectural backbone predominantly relies on either a Transformer \citep{NIPS2017_3f5ee243} or Conformer \citep{gulati20_interspeech} encoder. Second, the diversity extends to the training data, which are not public for most SFMs, except for wav2vec 2.0 and NeMo STT Fast Conformer. Third, distinctions emerge in the supported languages, as most SFMs are limited to English, while the Whisper encoder supports
99 languages \citep{pmlr-v202-radford23a}.
Lastly, SFMs vary in the tasks they undertake, with some focusing solely on ASR, while
Whisper extends its capabilities to ST and timestamp prediction.
In addition, it is noteworthy that the majority of the SFMs used are not publicly available: none of the \sara{four} works that trained a custom speech model released it, and USM, employed by~\pfive, is not openly accessible. From these observations, it is evident that the works are not directly comparable, and is often impossible for future research to make fair comparisons with existing solutions. The absence of a comparative analysis among SFMs also hinders our understanding of their impact on downstream performance, as well as the identification of the most suitable choice to guide future research.

\paragraph{Length Adapter (LA).} This module is designed to reduce the number of embeddings representing an audio sequence over the time axis. This operation serves a dual 
purpose. 
\sara{On the one hand, compressing the length of audio sequences 
\lb{--}
typically longer than the corresponding textual ones
\lb{--}
contributes to reducing the difference between the two modalities, hence limiting the modality mismatch for the LLM, which is trained on textual inputs.}
On the other,
as 
\mg{current}
LLMs exploit the Transformer architecture whose self-attention suffers from a quadratic complexity with respect to the input sequence, this compression prevents the already demanding memory and computational costs from becoming unaffordable\mg{.}
As already noted for the SFM, Table \ref{tab:speech_llm_components} highlights that 
a wide range of methods have been adopted for the LA. 
Also in this case, a comparison between different solutions in the same settings is missing, with one exception. In fact, \pthree{} evaluates two LA methods based on a CTC module \citep{Graves2006ConnectionistTC}: \textit{i)} the CTC compression, which averages vectors corresponding to the same CTC predictions, and \textit{ii)} the CTC blank filtering \citep{10096065},
which discards all the vectors corresponding to predictions of the \texttt{<blank>} token\mg{.}\footnote{\mg{\texttt{<blank>} is a special token used by the CTC loss to denote the absence of speech content in the signal (e.g., silence).}}
Their results indicate that the former leads to better ST quality.
The only other existing comparison of LAs has been conducted 
\mg{in the scope of}
the \mg{related} ASR task, where
\citet{yu2023connecting} introduce a window-level Q-Former \citep{10.5555/3618408.3619222} encoder, named Seg-QF, demonstrating its superiority over a plain Q-Former, a 1D convolutional layer, and the stacking of consecutive vectors followed by a feed-forward network \mg{(as done in \pnine)}. Seg-QF is very similar to the Window-level Q-Former used in \psix: it divides the speech sequence into chunks of a predetermined size (a hyperparameter $n_s$) that are independently processed by the Q-Former, which controls the length of the output sequence with the number of learned query vectors used (another hyperparameter $n_q$). As a result, this approach reduces the input length by a factor of $n_s/n_q$. It is important to notice that this finding \sara{was} obtained by keeping both the SFM and the LLM frozen and without introducing any other module (e.g., without any modality adapter). Hence, its validity should be \sara{confirmed} in different conditions where the LA does not have to learn the modality mapping as well.
\mg{To sum up, although the literature offers insights into the most promising approaches for LAs, a comparative analysis covering all the proposed methods is missing. Moreover, as the LA controls the length of the LLM input and this is a critical factor for the computational costs of the resulting SFM+LLM models, their analysis should not be limited to the downstream (ST) performance but \mn{it should} also 
\mn{consider}
the model efficiency, which has been disregarded so far.}

\paragraph{Modality Adapter (MA).} 
\mn{The MA is a small trained network (compared to SFMs and LLMs) that maps the LA output into an embedding space compatible with the LLM.}
Compared to LA, \sara{its design} has seen fewer variations: in some instances, the MA is a simple FFN (\pone{} and \pnine) or is composed of a variable number of Transformer layers 
\mn{(\pthree, \pfour, and \pfive). In other cases,}
\mg{it is fused with the LA}
(\ptwo{} and \psix{}) or even absent (\pseven{} and \peight).
The necessity and design of the MA depend on the 
\mn{training strategy adopted:} 
\mg{if the LLM is finetuned, the MA can \mn{indeed} be avoided (see \pseven{} and \peight) as the LLM can learn to use a new embedding space 
\mn{(the one produced by the SFM and LA)}\mg{. In contrast,}
if the LLM and SFM are not adapted, the MA is \sara{necessary} to enable their communication (as in \pfive).}
Similarly, the complexity and size of the MA \sara{can} vary depending on the training strategy: \sara{if a simple MA is adopted, the introduction of trainable adapters in the LLM or its finetuning may be required (\pone{} and \ptwo).}
\mn{However, the role and necessity of the MA have not been systematically investigated in existing works, which introduced it without \sara{conducting} ablation studies or analyses on its size. This calls for a dedicated contrastive evaluation \sara{accounting} for crucial factors like the training strategy and the quantity of finetuning paired data used.}

\paragraph{Prompt-Speech Mixer (PSMix).} 
The goal of the PSMix is to merge the speech representation with the textual prompt that is to be fed to the LLM. 
Regarding the type of textual prompt, 
\lb{the analyzed works show little variability,}
with most of them relying on a fixed template to fill with the source and target language (e.g. ``\textit{Translate the audio from <SOURCE LANGUAGE> to <TARGET LANGUAGE>}''). In \pthree{}, the authors experimented with 
a list of templates \sara{to enhance system robustness}, but they did not investigate \sara{its impact on} 
performance. In \pfive{}, the authors demonstrated that a wider range of prompts 
enables
the system to support
\mg{unseen}
ones at inference time; however, in their setting, this corresponds to a broader set of tasks\sara{, making it challenging to isolate}
the contribution of different prompts and tasks to this ability.
Regarding the PSMix strategies, most works rely on three concatenation solutions: prepending the speech representation to the prompt embeddings (\ptwo, \pfour, \psix, \pseven{} and \peight), 
\mn{appending it}
to the prompt embeddings (\pthree{} and \pfive{}), or interleaving the 
speech representation with a prompt prefix and suffix (\pnine). 
Only one work (\pone) completely omits the prompt
\mg{and the PSMix module by directly feeding the LLM with the speech representations.}
To sum up, it is unclear whether using a fixed template for the prompt is the best choice, despite its prominent adoption, and which of the PSMix options (if any) leads to the best results. As these aspects have not yet been thoroughly studied, such interesting questions remain to be addressed in future works.


\paragraph{LLM.} The last component is the LLM, which takes the mixed prompt and speech representations as input  
\mn{to generate the final}
(textual) translation. In \pfive, \citet{wang2023slm} claim that ``the pretrained LLM plays a crucial role in both training efficiency and model quality'', and that a stronger model on a given task  leads to better performance. 
\lb{However,} with the only exception of works by the same authors (\pthree{} and \pfour) that leverage LLaMa2 7B, all the SFM+LLM combinations 
exploit
different LLMs
\mg{without motivating the choice (e.g. through comparisons across models):}
\pone{} uses a larger LLaMa2 (i.e., the 13B version), \psix{} and \pnine{} use a finetuned version 
\mg{(}Vicuna 13B and LLaMa2 Chat 7B, respectively) while \ptwo{}, \pfive{}, \pseven{}, and \peight{} use completely different models.
The 
dominance
of the LLaMa family 
is probably motivated by
its openness and support for multiple languages. On the other hand, LLMs specifically built for the translation task are emerging \citep{xu2023paradigm} and represent a natural option
\mg{to be considered in future works.}
In light of
the high computing costs of these large models and the significant 
performance variations
they can exhibit, establishing the best option for the ST task through systematic comparisons represents 
\mg{a}
priority
for future research.

\begin{table*}[!t]
\small
    \centering
        \begin{tblr}{
      colspec={|X[0.1,c]|X[0.7,c]|X[1.4]|X[1.1]|X[0.3,c]|X[0.4,c]|X|X|}, row{1} = {c}, hlines,
    }
        \# & \textbf{Model} & \textbf{Train. Data} & \textbf{Train. Tasks} & \textbf{SFM fn} & \textbf{LLM fn} & \textbf{Eval. Data} & \textbf{Supported Lang. Pairs}  \\
        \hline
        \pone & LST & MuST-C, LibriSpeech & ASR, ST & No & Yes & MuST-C & en$\rightarrow$\{de, fr, es\} \\
        \ptwo & SALM & IWSLT 2023 & ASR, ST & No & LoRA & MuST-C & en$\rightarrow$\{de, ja\} \\
        \pthree & Speech- LLaMa & in-house & ASR, ST & \SetCell[r=2]{c} Yes & \SetCell[r=2]{c} LoRA & CoVoST2 & \{de, zh, ar, es, fr, it, nl, ja, ru, pt, et, sv, sl\}$\rightarrow$en \\
        \pfour & COSMIC & TEDLIUM3 & ASR, SQA & & & TEDLIUM3, FLEURS & en$\rightarrow$\{es, fr, de, zh\} \\
        \pfive & SLM & Alpaca, CoVoST2, YouTube (in-house) & ASR, ST, SIT & No & No & CoVoST2 & \{fr, de, es, ca, it, ru, pt, fa, et, mn, nl, tr, ar, sv, lv, sl, ta, ja, id, cy, zh\}$\rightarrow$en \\
        \psix & SALMONN & AudioCaps, Clotho, CoVoST2, GigaSpeech, IEMOCAP, LibriMix, LibriSpeech, MillionSong, MusicCaps, MusicNet, VoxCeleb1, WavCaps & ASR, ST, AAC, PR, ER, MC, OSR, SV, GR, SQA, AQA, MQA, ABST & No & LoRA & CoVoST2 & en$\rightarrow$\{de, ja, zh\} \\
        \pseven & LLM-ST & CoVoST2, GigaST, MuST-C v2, WeNetSpeech + in-house & ASR, ST, MT, PT, ITN, TE, SRST, STST & Yes & Yes & CoVoST2, GigaST, MuST-C v2, in-house & en$\leftrightarrow$zh \\
        \peight & Qwen-Audio & in-house & ASR, ST, OSR, DASR, SRWT, DID, LID, GR, ER, SV, SD, SER, KS, IC, SF, SAP, VSC, AAC, SEC, ASC, SED, AQA, SID, MC, MIC, MNA, MR, MQA & Yes & No & CoVoST2 & en$\rightarrow$\{de, zh\}, \{de, zh, es, fr, it\}$\rightarrow$en \\
        \pnine & Conformer LLaMa & MLS & ASR & Yes & No & N/A & N/A \\
        \end{tblr}
    \caption{Experimental settings adopted for finetuning the SFMs+LLMs. "fn" stands for finetuning, and, for supported language pairs (Supported Lang. Pairs), we intend language pairs on which models have been evaluated.}
    \label{tab:speech_llm_settings}
\end{table*}

\subsection{Training and Evaluation}
\label{subsec:train-eval}

In this section, we describe the experimental settings of the analyzed papers by focusing on the datasets used for training and evaluation, the supported tasks and language pairs, and the techniques used 
for SFM and LLM finetuning.
A summary is provided in Table \ref{tab:speech_llm_settings}.
\mg{The task acronyms are defined in Appendix \ref{app:acronyms}, the training and evaluation datasets are 
\mn{reported}
in Appendix \ref{app:datasets}, while the language codes follow the ISO 639 notation.\footnote{\url{https://www.iso.org/standard/74575.html}}}

\paragraph{Training Data.}
The training datasets 
\mn{used in}
the 9 analyzed papers are different both in terms of type and quantity. Approximately half of the 
\mg{works}
(5 out of 9)
\mg{leverage}
publicly available data only, both within and outside the ST domain. Despite this, none of them utilize similar data settings for finetuning their proposed SFM+LLM architecture: while
LST \pone, COSMIC \pfour, and ConformerLLaMa \pnine{} rely on 1 or 2 datasets, SALM \ptwo{} 
\mg{uses}
all the 11 
\mn{speech corpora available}
for the IWSLT 2023 Offline Speech Translation Shared 
Task,\footnote{\url{https://iwslt.org/2023/offline}} and SALMONN \psix{} employs 12 different datasets during training.
For SFM+LLM models trained on non-publicly available data, we observe that SLM \pfive{} and LLM-ST \pseven{} adopt a combination of in-house and open data, while Speech-LLaMa \pthree{} and Qwen-Audio \peight{} exclusively use proprietary data. 
In addition to the lack of uniform training settings, 
none of the existing works has analyzed scaling laws and the effect of increasing the data size on the performance, rendering a fair comparison among the diverse approaches impractical.


\paragraph{Training Tasks.}
Regarding training tasks, almost half of the SFM+LLM models (5 out of 9) extend their scope beyond pure ASR and ST applications. Among them, SLM \pfive{} integrates a single additional task -- instruction tuning --
while
LLM-ST \pseven{} is trained with 4 translation-related tasks (e.g., translation explanation) and 2 speech-related tasks (e.g., timestamp 
\mg{estimation).}
In contrast, SALMONN \psix{} supports a diverse array of up to 10 additional tasks, spanning various domains such as
\mg{SQA}
and emotion recognition. Qwen-Audio \peight{} takes this a step further by incorporating 26 more tasks, encompassing a comprehensive collection of speech, audio, and music-related tasks.
In contrast, COSMIC \pfour{} is exclusively trained on ASR and
\mg{SQA}
but is also tested on ST. Similarly, ConformerLLaMA \pnine{} is trained solely on the ASR task but it demonstrates emergent capabilities in ST, although its translation quality is not systematically assessed.\footnote{\label{foot:anec}\mg{The ST ability \sara{of the model} is only anecdotally reported.}}
Interestingly, only three models -- LST \pone, SALM \ptwo, and Speech-LLaMa \pthree{} -- are trained on the same tasks (ASR and ST)\mg{.}
\mg{The effect of adding more tasks on the resulting model capabilities and ST performance is (partly) studied only in SALMONN \psix, where tasks are progressively introduced.}
Specifically, its training strategy involves three stages: \textit{i)} the first stage (pre-training) includes ASR and AAC, \textit{ii)} the second stage (instruction tuning) includes 12 tasks, and \textit{iii)} the third stage (activation tuning) finetunes the model on tasks with longer and more diverse responses as \sara{AQA} and \sara{ABST}. The last step is shown to increase the generalization and emergent abilities while impacting translation quality in 
\mg{a limited yet unclear}
way, as it improves 
\lb{in one direction \mg{(en-ja)} but degrades in two other directions \mg{(en-de and en-zh)}.}
All in all, the lack of uniformity in the training task selection hinders the comparability of the solutions, and the benefits of knowledge transfer across tasks \citep{knowledgetransfer,ke2021achieving,kubo2022knowledge} have yet to be studied in-depth.

\paragraph{SFM and LLM finetuning.}
As SFMs and LLMs are huge in terms of parameters, their training/finetuning is computationally expensive. This raises the question 
about
whether they can be used without expensive adaptation or not. \sara{Regarding the SFM, more than half of the examined papers (5 out of 9) finetune it, while this component is kept frozen in the others}. The LLM, instead, is adapted by 6 of the 9 analyzed papers, but only 2 (\pone{} and \pseven) finetune the whole model. 
The others rely on the Low-Rank Adaptation \citep{hu2022lora}\mg{, or LoRA, a widely employed technique for adapting LLMs to new datasets or tasks \citep{hu2023bliva,anonymous2024datainf}. LoRA}
consists in introducing trainable rank decomposition matrices into each layer of the architecture while keeping the original weights frozen, so as to significantly reduce the trainable parameters (by a factor of 10,000). 
\lb{Notably, only one study (\pfive) presents results with both the SFM and LLM frozen, and also shows that LLM finetuning yields substantial performance gains. However, since finetuning is  conducted on data from the same domain as the test set, the observed benefits may be partially attributed to domain adaptation, making it challenging to quantify the improvement solely attributable to finetuning.}
\lb{Similarly, \citet{wu2023decoder} (\pthree) show that LoRA leads to improvements of $\sim$1.5 BLEU, averaged over 13 CoVoST2 language pairs.} 
We can conclude that, while LLM adaptation brings significant improvements, it is unclear whether the need for finetuning depends on the type of LLM used (e.g., would it be needed when using an LLM built for the translation task?) or on the design of other modules (e.g., the MA) or on other training choices (e.g., adapting the SFM or not). \sara{Moreover, similar}
studies should
be conducted for the even less explored SFM adaptation.

\paragraph{Evaluation Data.}
The selection of consistent evaluation benchmarks is crucial for facilitating meaningful comparisons among different 
SFM+LLM models. 
\mn{However, our survey reveals disparate choices regarding the test sets employed.} 
\mn{The main}
dichotomy regards the evaluation 
within
English-to-many or 
many-to-English 
settings,
as four papers focus on the former, two on the latter, \sara{\mg{and} two on both (although \pseven{} 
\lb{investigates only} zh), while one (\pnine{}) does not report evaluation results.}\textsuperscript{\ref{foot:anec}}
CoVoST2 emerges as the most widespread benchmark (used in 5 papers), thanks to its broad coverage of translation directions (15 in the English-to-many case, and 21 in the many-to-English one). For the English-to-many scenario, MuST-C is also frequently used (in 3 cases), while COSMIC \pfour{} is the only \sara{one} tested on TEDLIUM and FLEURS, and LLM-ST \pseven{} complements CoVoST2 and MuST-C with GigaST and private in-house test sets. 
The tendency not to report scores computed on a common set of benchmarks and
\lb{language pairs (as discussed below),}
contributes to making the comparison for future works nearly impossible without an expensive re-implementation of existing methods, slowing down the progress in the area.


\paragraph{Supported Translation Languages.}
Concerning the languages supported for translation, all the examined papers analyze different pairs but share 
\mg{the}
characteristic of being English-centric. They investigate either many-to-English directions (\pthree, \pfive, and \peight) or English-to-many directions (\pone, \ptwo, \pfour, \psix, and \peight). In the context of many-to-English pairs, Qwen-Audio \peight{} encompasses 5 source languages, Speech-LLaMa \pthree{} covers more than half of the CoVoST2 languages (13 out of 21), while SLM \pfive{} includes all 21 CoVoST2 pairs.
Conversely, all papers focusing on English-to-many directions cover 2 to 4 target languages, constituting a consistently smaller set compared to the many-to-English case. \mg{Lastly,}
LLM-ST \pseven{}
exclusively addresses a single translation pair (en$\leftrightarrow$zh).
Interestingly, the majority of the works \sara{mainly} report results for 
either de$\rightarrow$en or en$\rightarrow$de, which represents one of the most extensively analyzed language 
\mn{pairs}
in ST \citep{anastasopoulos-etal-2021-findings,anastasopoulos-etal-2022-findings,agrawal-etal-2023-findings}\mg{, with \pseven{} being the only work addressing neither of them.}
\mg{en$\leftrightarrow$zh}
emerges as the second most reported language 
\mn{setting}
\mg{(each direction being used by 4 papers).}
Despite these commonalities, it is evident that the choice of supported languages varies significantly between the works\mg{. Also,}
the impact on 
performance of supporting multiple languages -- which can interfere or enable transfer learning between linguistically similar languages
\citep{ruder-etal-2019-transfer,durrani-etal-2021-transfer}
-- remains uncertain.

\section{What is Missing?}
\label{sec:missing}

\mn{Alongside the need for focused and thorough analyses devoted to identifying the best-performing option for each architectural building block highlighted in \S\ref{subsec:architecture} and the effects of the training choices discussed in \S\ref{subsec:train-eval}, in the following we identify blind spots that need to be addressed for a more grounded and insightful progress in research on SFM+LLM solutions for ST.}

\paragraph{Open Standard Training Settings.} As highlighted 
throughout the previous section, the lack of common experimental settings prevents the fair and direct comparison of different works. The 
\mg{adoption}
of public and standard
\mg{training}
settings holds paramount importance in advancing research and fostering progress within the scientific community \citep{koch2021reduced}\mg{. On the one \sara{hand}, it enables}
the comparison among various works,
\lb{thus providing}
\mn{actionable}
insights on the most promising architectural choices. 
\mg{On the other, it}
\sara{fosters}
inclusivity and accessibility, 
\mg{allowing researchers 
\mn{without access to large proprietary corpora}
to contribute to the field}
\citep{scandura2020academic,10.1093/scipol/scab010}, and \lb{thus} supporting AI democratization \sara{in the development process} \citep{seger2023democratising}.
Therefore, we advocate for future research to adhere to established data-setting standards,
paving the way for cumulative progress and shared 
understanding 
\sara{in} the field.
\mg{However, as}
experimenting with different data sizes is also an interesting topic and findings may vary depending on the datasets and the tasks used in the training stage (see \S\ref{subsec:train-eval}), it is debatable which would be the most appropriate training set. 
\mg{In}
the English-to-many scenario, researchers commonly 
\mn{adhere to}
the IWSLT offline
\mg{constrained \mn{data} condition,}\footnote{\url{https://iwslt.org/2023/offline}} comprising $\sim$4.5K hours of English audio, while, for smaller-scale experiments, MuST-C ($\sim$500 hours) is a widespread option.
\mn{For many-to-English settings,}
CoVoST2, mTEDX \citep{salesky21_interspeech}, and Europarl-ST \citep{9054626} are open datasets with ST references 
and
can be complemented with larger ASR resources such as CommonVoice \citep{ardila-etal-2020-common} and VoxPopuli. Notice that, by advocating for standardized and public training data, we do not imply that researchers should not investigate the effects of training in different data conditions. Rather, we suggest that, for works primarily focused on defining new architectural solutions, reporting results for (at least) 
\mg{a standard setting}
would ease 
\mn{comparisons}
with other alternatives and reduce the overall computational costs.

\paragraph{\mg{Standard and Reliable} Evaluation.}
The comparison between different methods is currently hindered not only by different training conditions but also by 
\label{}{the fact that practitioners do not systematically present results  on a common open benchmark.}
Furthermore,
all works rely on the BLEU metric \citep{papineni-etal-2002-bleu}, except for \pseven, which additionally reports COMET \citep{rei-etal-2022-comet}.
Although we acknowledge that BLEU is still widespread \citep{marie-etal-2021-scientific} despite
the wide consensus on 
its limited dependability 
and correlation with human judgments
\citep{freitag-etal-2022-results}, 
we argue that this specific scenario exacerbates the need for adopting alternative metrics to assess translation quality. The main reason behind this argument is the well-known tendency of n-gram-based metrics to penalize translations generated by LLMs that are, in general, less literal \citep{LLMSurvey,liu-etal-2023-g}.
As a suggestion for future works, we recommend reporting at least one semantic metric (e.g., COMET), and, preferably, multiple metrics. We also believe that reporting scores on open and multilingual benchmarks, such as CoVoST2, would improve comparability across studies without the need for re-running costly experiments, 
thereby promoting faster, cost-effective progress within the research community.

\paragraph{Comparison with Standard ST Approaches.} In analogy to studies \citep{sperber-paulik-2020-speech,bentivogli-etal-2021-cascade} and initiatives \citep{agrawal-etal-2023-findings} dedicated to assessing the strengths and weaknesses \lb{of the 
two established
end-to-end and cascade ST paradigms,}
 \mn{the emergence of the SFM+LLM solution}
calls for 
\mn{thorough and fine-grained evaluations}
to investigate its peculiarities compared to \lb{other, more traditional methods}. 
\mn{This need is also motivated by a}
recent analysis in the context of text-to-text translation \citep{pang2024salute}, which \mg{showed} that LLMs are partly affected by
\mn{long-standing} problems of neural approaches (e.g., the translation of rare entities and out-of-domain settings), while they do not face others (e.g., \mn{the} translation of long sentences) and 
\mg{suffer from new ones}
(e.g., pre-training data imbalance across domains and languages).
Among the new problems, a noteworthy element is the inference efficiency: the comparison with the standard methods -- which typically rely on models of limited size (100-300M parameters) -- should account for this aspect, which is critical for social, economic, and environmental reasons \citep{strubell-etal-2019-energy}. Along this line, important research directions include \textit{i)} pruning the LLM (and possibly the SFM) in a task-aware manner \citep{ma2023llmpruner,zhu2023survey,dery2024everybody}, \textit{ii)} dynamic layer selection during decoding \citep{xin-etal-2020-deebert,geva-etal-2022-transformer,xia2024unlocking}, and \textit{iii)} efficient decoding strategies \citep{10.5555/3327546.3327673,chen2023accelerating,10.5555/3618408.3619203,santilli-etal-2023-accelerating}.
In addition, the speech source 
\mg{
\lb{contains a wide}  range of information 
\lb{that can be exploited depending }
on the paradigm used (e.g., prosody is not handled by cascade systems -- \citealt{zhou2024prosody}). As such, the ability of SFM+LLM models to leverage 
\lb{this information}
has to be investigated.}
The fine-grained evaluation of these aspects calls for the comparison of SFM+LLM models with other paradigms on tailored test suites \citep{king-falkedal-1990-using,ribeiro-etal-2020-beyond}, similar to those used in MT \citep{kocmi-etal-2023-findings}.

\paragraph{In-Context Learning \mg{Assessment}.} One of the most interesting emergent abilities of 
LLMs \citep{wei2022emergent} is 
\lb{their ability} to exploit a few demonstrations or examples to perform a task or \sara{enhance their} performance \mg{on it} \citep{dong2023survey}. This ability -- referred to as in-context learning (ICL) 
\citep{NEURIPS2020_1457c0d6} -- 
\mg{is one of the main motivations for integrating an SFM and an LLM into 
\mn{a single}
ST model.}
\mn{However, the transfer of ICL capabilities of LLMs to the speech modality, and, even more so, to the SFM+LLM approach to ST, cannot be taken for granted.}
\mg{In fact, while the ICL ability of SFM+LLM has been successfully assessed in ASR and \sara{SLU}}
\citep{gao22e_interspeech,hsu2023exploration,chen2023salm} -- also within the retrieval-augmented framework \citep{wang2023speechtotext}, where the relevant context is retrieved from a knowledge base \citep{ram-etal-2023-context} -- the only attempt in ST \mg{has not been similarly successful} \citep{chen2023salm}. Moreover, SFMs like Whisper 
\mn{feature}
similar \mg{(yet limited)} \lb{ICL} capabilities \citep{peng23d_interspeech,wang2023whisper}, which might make the SFM+LLM integration not even necessary. For this reason, investigating 
whether and 
\mn{to what extent}
the integration of SFMs with LLMs 
\mn{transfers}
\mg{the ICL ability \sara{of the latter} to the}
ST task is an important and interesting avenue for future studies.

\section{Conclusions}

\mg{
The ST landscape has recently witnessed the emergence of a new paradigm, which is the combination of SFMs and LLMs into single ST models. 
To summarize the lessons learned and \sara{establish a unified framework},
we surveyed the existing work\lb{s} on the topic, analyzing their
architectural
and training 
choices.
As a result, we identified
a common abstraction of the surveyed SFM+LLM architectures, which consists of five building blocks:
\textit{i)} the SFM extracting high-level speech representations, \textit{ii)} the Length Adapter compressing such sequence of features, \textit{iii)} the Modality Adapter mapping them to an embedding space more suitable for the LLM, \textit{iv)} the Prompt-Speech Merger combining the speech information with an adequate prompt for the LLM, and \textit{v)} the LLM generating the output translation. Subsequently, we highlighted how the current lack of standardized training recipes and evaluations hinders the direct comparison of the proposed approaches, 
limiting the possibility of extracting precise and unified 
indications.
Lastly, we pointed out the need for thorough comparisons with standard ST approaches and in-depth investigations of the inherent capabilities of
SFM+LLM solutions in order to shed light on its real potential for ST.}



\section*{Limitations}

Our survey of the existing studies on the integration of an SFM and an LLM has been limited to the context of the speech-to-text translation task. We did not target the more generic case of the SFM+LLM integration as already covered by existing surveys \citep{latif2023sparks} and would have prevented the ability to go more in-depth for the specific works within the page limit.
For \sara{the same reason}, we have not included works that target different tasks, such as ASR
\citep{chen2023x,hono2023integration,lakomkin2023endtoend,radhakrishnan-etal-2023-whispering,yu2023connecting}, nor models that focus on audio phenomena different from the human speech\footnote{E.g., sound classification/captioning or music processing.} 
\citep{deshmukh2023pengi,han2023onellm,shu2023llasm,zhang2023metatransformer,zhao2023chatbridge}. While they can inspire effective solutions for the ST case as well, assessing their portability to the ST field may be the focus of dedicated works.

Moreover, we only discussed the solutions in terms of their ST performance, without considering their generalization capability and/or capacity to perform different downstream tasks, as it would have added complexity to an analysis that targeted a specific task of spoken language processing.
However, we believe that \sara{applying foundation models to a specific task} does not necessarily imply that they need to retain generic capabilities, although this is a desirable property. Similarly, we have not delved into ethical considerations and implications of such solutions \citep{manvi2024large,Schramowski2022}, as we believe that it should be the topic of tailored and dedicated evaluations, also in comparison with traditional ST approaches, as mentioned in \S\ref{sec:missing}.

Lastly, the study did not include models that can perform the ST task as part of a cascade approach, where audio is converted into text or other units \citep{wang2023viola,zhang-etal-2023-speechgpt}, nor those that use the LLM only to understand user requests and forward their actual processing to SFMs \citep{huang2023audiogpt}. While these \sara{represent} viable solutions, we argue that their progress and analysis are directly linked to the ASR quality of SFMs and the MT quality of LLMs, which are extensively studied in specific works \citep{pmlr-v202-radford23a,hendy2023good,communication2023seamlessm4t,xu2023paradigm,pang2024salute}.

\section*{Acknowledgements}
We acknowledge the support  of the PNRR project FAIR -  Future AI Research (PE00000013),  under the NRRP MUR program funded by the NextGenerationEU. This paper has received funding from the European Union’s Horizon research and innovation programme under grant agreement No 101135798, project Meetween (My Personal AI Mediator for Virtual MEETtings BetWEEN People).

\bibliography{custom}

\newpage

\appendix

\section{Task Acronyms}
\label{app:acronyms}

Table \ref{tab:tasks} report the list of tasks acronyms used in \S\ref{subsec:train-eval}.

\begin{table}[!ht]
\small
    \centering
        \begin{tblr}{
      colspec={|X[0.2]|X|}, row{1} = {c}, hlines,
    }
        \textbf{Acronym} & \textbf{Full Task Name} \\
        \hline
        AAC & Automatic Audio Captioning \\
        ABST & Audio-based Storytelling \\
        AQA & Audio Question Answering \\
        ASC & Acoustic Scene Classification \\
        ASR & Automatic Speech Recognition \\
        DASR & Automatic Dialect Speech Recognition \\
        DID & Dialect Identification \\
        ER & Emotion Recognition \\
        GR & Gender Recognition \\
        IC & Intent Classification \\
        ITN & Inverse Text Normalization \\
        KS & Keyword Spotting \\
        LID & (spoken) Language Identification \\
        MC & Music Captioning \\
        MIC & Music Instruments Classification \\
        MNA & Music Note Analysis (e.g. pitch, velocity) \\
        MQA & Music Question Answering \\
        MR & Music Recognition (including genre) \\
        MT & Machine Translation \\
        OSR & Overlapped Speech Recognition \\
        PR & Phone Recognition \\
        PT & Pronunciation Translation \\
        SAP & Speaker Age Prediction \\ 
        SD & Speaker Diarization \\
        SEC & Sound Event Classification \\
        SED & Sound Event Detection \\
        SER & Speech Entity Recognition \\
        SID & Singer Identification \\
        SF & Slot Filling \\
        SIT & Speech Instruction Tuning \\
        SQA & Speech/Spoken Question Answering \\
        SLU & Spoken Language Understanding \\
        SRST & Speech Recognition with Sentence-level Timestamps \\
        SRWT & Speech Recognition with Word-level Timestamps \\
        ST & Speech Translation \\
        STST & Speech Translation with Sentence-level Timestamps \\
        SV & Speaker Verification \\
        TE & Translation Explanation \\
        VSC & Vocal Sound Classification \\
        \end{tblr}
    \caption{List of tasks with their acronyms.}
    \label{tab:tasks}
\end{table}

\section{List of Datasets}
\label{app:datasets}
Table \ref{tab:datasets} lists the dataset mentioned in \S\ref{subsec:train-eval}, with their reference, and the indication of whether they are open and contain ST references.

\begin{table*}[!ht]
\small
    \centering
        \begin{tblr}{
      colspec={|X[0.1,c]|X[4]|X[3]|X[c]|X[c]|}, row{1} = {c}, hlines,
    }
        \# & \textbf{Name} & \textbf{Paper/Reference} & \textbf{Open} & \textbf{ST} \\
        \hline
        1 & MuST-C & \citet{di-gangi-etal-2019-must} & \yes & \yes \\
        2 & LibriSpeech & \citet{7178964} & \yes & \no \\
        3 & IWSLT 2023 Offline Speech Translation Shared Task & \citet{agrawal-etal-2023-findings} & \yes & \yes \\
        4 & TEDLIUM3 & \citet{hernandez2018ted} & \yes & \no \\
        5 & GigaSpeech & \citet{chen21o_interspeech} & \yes & \no \\
        6 & AudioCaps & \citet{kim-etal-2019-audiocaps} & \yes & \no \\
        7 & Clotho & \citet{9052990} & \yes & \no \\
        8 & IEMOCAP & \citet{Busso2008IEMOCAPIE} & \yes & \no \\
        9 & MusicCaps & \citet{agostinelli2023musiclm} & \yes & \no \\
        10 & LibriMix & \citet{cosentino2020librimix} & \yes & \no \\
        11 & VoxCeleb1 & \citet{NAGRANI2020101027} & \yes & \no \\
        12 & MillionSong & \citet{Bertin-Mahieux2011} & \yes & \no \\
        13 & MusicNet & \citet{thickstun2017learning} & \yes & \no \\
        14 & MLS (Multilingual LibriSpeech) & \citet{pratap20_interspeech} & \yes & \no \\
        15 & Alpaca & \citet{alpaca} & \yes & \no \\
        16 & CoVoST2 & \citet{wang21s_interspeech} & \yes & \yes \\
        17 & YouTube & \citet{zhang2023google} & \no & \no \\
        18 & CoVoST2 & \citet{wang21s_interspeech} & \yes & \yes \\
        19 & GigaST & \citet{ye23b_interspeech} & \yes & \yes \\
        20 & MuST-C v2 & \citet{CATTONI2021101155} & \yes & \yes \\
        21 & WeNetSpeech & \citet{9746682} & \yes & \no \\
        22 & FLEURS & \citet{10023141} & \yes & \yes \\
        23 & SpeechStew & \citet{chan2021speechstew} & \yes & \no \\
        24 & VoxPopuli & \citet{wang-etal-2021-voxpopuli} & \yes & \yes \\
        25 & Multi-context TTS & \citet{10023323} & \yes & \no \\
        26 & Inspec & \citet{hulth-2003-improved} & \yes & \no \\
        27 & WikiQA & \citet{yang-etal-2015-wikiqa} & \yes & \no \\
        28 & SLURP & \citet{bastianelli-etal-2020-slurp} & \yes & \no \\
        29 & AISHELL-1 & \citet{aishell_2017} & \yes & \no \\
        30 & AISHELL-2 & \citet{du2018aishell2}  & \yes & \no \\
        31 & Industrial Data & \citet{gao23g_interspeech} & \yes & \no \\
        32 & CochlScene & \citet{jeong2022cochlscene}  & \yes & \no \\
        33 & TUT2017 & \citet{Mesaros2016_EUSIPCO}  & \yes & \no \\
        34 & MELD & \citet{poria-etal-2019-meld}  & \yes & \no \\
        35 & ClothoAQA & \citet{lipping2022clotho}  & \yes & \no \\
        36 & VocalSound & \citet{9746828}  & \yes & \no \\
        37 & NSynth & \citet{10.5555/3305381.3305492}  & \yes & \no \\
        \end{tblr}
    \caption{Datasets used in the surveyed papers and whether they are open (Open) and contain ST references (ST).}
    \label{tab:datasets}
\end{table*}

\end{document}